\documentclass{article}
\usepackage{spconf,amsmath,graphicx}
\usepackage[bottom]{footmisc}
\usepackage{hyperref}
\usepackage{multirow}
\usepackage{booktabs}
\usepackage{makecell}
\usepackage{pgfplots}
\pgfplotsset{compat=newest}
\usepackage{caption	}
\usepackage{siunitx}

\title{AUDIOVISUAL HIGHLIGHT DETECTION IN VIDEOS\sthanks{Paper accepted at ICASSP 2021.}}
\name{
Karel Mundnich$^{\star}$\sthanks{Work performed during an internship at Amazon.}\qquad
Alexandra Fenster$^{\ddagger}$\qquad
Aparna Khare$^{\ddagger}$\qquad
Shiva Sundaram$^{\ddagger}$
}
\address{
\small{$^{\star}$Signal Analysis and Interpretation Lab, University of Southern California, Los Angeles, CA 90089}\\
\small{$^{\ddagger}$Amazon Lab126, Sunnyvale, CA 90489}
}
\newcommand{\alignedsection}[1]{
	\vspace{-3mm}
	\section{#1}
	\vspace{-3mm}
}

\newcommand{\alignedsubsection}[1]{
	\vspace{-4mm}
	\subsection{#1}
	\vspace{-2mm}
}

\newcommand{\alignedsubsubsection}[1]{
	\vspace{-3mm}
	\subsubsection{#1}
	\vspace{-2mm}
}

\newcommand{\alignedsectionsubsection}[2]{
	\vspace{-3mm}
	\section{#1}
	\vspace{-2mm}
	\subsection{#2}
	\vspace{-2mm}
}

\newcommand{\alignedsubsectionsubsubsection}[2]{
	\vspace{-3mm}
	\subsection{#1}
	\vspace{-1mm}
	\subsubsection{#2}
	\vspace{-1mm}
}

\begin{document}
\ninept
\renewcommand{\baselinestretch}{0.9}
\setlength{\skip\footins}{6pt}
\maketitle
\begin{abstract}
In this paper, we test the hypothesis that interesting events in unstructured videos are inherently audiovisual.  We combine deep image representations for object recognition and scene understanding with representations from an audiovisual affect recognition model. To this set, we include content agnostic audio-visual synchrony representations and mel-frequency cepstral coefficients to capture other intrinsic properties of audio.  These features are used in a modular supervised model. We present results from two experiments: efficacy study of single features on the task, and an ablation study where we leave one feature out at a time. For the video summarization task, our results indicate that the visual features carry most information, and including audiovisual features improves over visual-only information. To better study the task of highlight detection, we run a pilot experiment with highlights annotations for a small subset of video clips and fine-tune our best model on it. Results indicate that we can transfer knowledge from the video summarization task to a model trained specifically for the task of highlight detection.
\end{abstract}

\alignedsection{Introduction}
Finding highlights or \textit{interesting} segments in audiovisual content has become an increasingly important task in video processing. With the proliferation of smartphones and widely accessible video cameras, the amount of audiovisual content and the demand to share them has increased over the years. This has also increased the need to develop automated tools to find interesting segments while filtering out portions that are redundant.\ This is a non-trivial task in an unconstrained setting as finding relevant portions (rather, what we define as relevant or interesting) of audiovisual content depends on the context, nature of the videos (i.e.\ moving or static, egocentric point of view vs. not \cite{gygli2014creating}) and topic or genre.

To this end, the problem of finding highlights in audiovisual content has been studied in different settings.\ For example, Wrede et al.\ \cite{wrede2003spotting} define hotspots in meetings as ``regions of meetings that are marked by high participant involvement, as judged by human annotators'',  while Lai et al.\ in \cite{lai2013detecting} define summarization hotspots as ``regions chosen for inclusion in a summary by human annotators''.\ These definitions capture the intrinsic subjectivity of the task: both are defined by the judgment of human annotators.\ These definitions also partially overlap with the definition of video summarization: ``[...] to generate a short summary of the content of a longer video document by selecting and presenting the most informative or interesting materials for potential users.'' as defined by Ngo et al. in \cite{ngo2009}. However, the goals of both tasks are different: for highlight detection, we try to find those segments of audiovisual content that are most \textit{interesting} (i.e., the best moments), while the goal for video summarization is to generate an overview of the content \cite{garcia2018phd,truong2007video}.

\alignedsubsection{Related work}
\label{ssec:literature}
A plethora of works have addressed both problems of video summarization and highlight detection. One of the first works in video summarization is by Smith et al.\ in \cite{smith1997video}, where the authors combine video features and language content to create video skims. In recent years however, visual features have been the focus for video summarization. For example, in \cite{gygli2014creating}, Gygli et al.\ use aesthetics/quality in frames, presence of landmarks, and face/person detection as features for summarization. Other approaches use visual features only \cite{zhang2016video,zhou2018deep, fajtl2018summarizing}.\ A few have also used more than one modality. For example, Song et al.\ in \cite{song2015tvsum} use the lexical information in video titles from YouTube to aid the summarization, while Jiang et al.\ \cite{jiang2019comprehensive} use video features as well as raw audio features for the same task.

For highlight detection, both visual-only and multimodal approaches exist. In \cite{yang2015unsupervised}, Yang et al.\ use visual 3D convolutional deep features and segments from edited videos to train an auto-encoder for highlight detection.\ Molino et al. in \cite{garcia2018phd} use GIF generation from \url{http://www.gifs.com} as labels to find personalized highlights in videos using only visual features. Both of these only use visual features from videos. In terms of multimodal processing, \cite{wan2004efficient,merler2018automatic,dagtas2004multimodal} have presented novel approaches to finding highlights in sport videos, while \cite{kostoulas2015dynamic} uses a multimodal approach for highlight detection in movies.\ These domains are however, constrained.\ In work not involving video, Makhervaks et al.\ in \cite{makhervaks2020combining} propose using acoustic information from speech, lexical content from transcripts, and interaction statistics of participants to find hotspots in meetings, and found that the lexical content was the most informative modality.

\vspace{-1mm}
\alignedsubsection{Contributions}
In contrast with the research discussed above, we take a modular supervised approach to the problem of highlight detection in unstructured and unconstrained audiovisual content. We combine representations from semantically meaningful models that take into account scene understanding, object recognition and face detection in vision, affect (arousal and valence states of individuals in the video) and synchrony in audiovisual content.\ We include audio-only features in the form of mel-frequency cepstral coefficients (MFCCs) that are popular in audio information retrieval problems.\ We study the performance of these modalities by training models over video summarization labels, whose predictions we use to create rankings for highlight detection.\ Moreover, we use these pre-trained models on in-house collected annotations over a small set of videos with social interactions, and fine-tune over this set of labels. We do so to explore the use of transfer learning from the video summarization task on a large unstructured dataset to the highlight detection task on a more specific domain as this technique has been successfully applied in both computer vision and speech. These visual, audiovisual and audio only representations stem from our motivation to analyze the relevance of audiovisual cues in summarization and video highlight detection tasks. While a truly end-to-end solution is our north star, we limit the scope of this study to analyze the performance of these modalities for our task.
\vspace{-1mm}
\alignedsectionsubsection{Methodology}{Dataset}
We use the CoView dataset \cite{CoView} that consists of 1500 videos. Each video in the train set contains summarization labels for each 5s-long segment, collected from 20 annotators, in the interval $[0,2]$ (where $0$ is not important at all and $2$ is very important). The train partition consists of 1200 annotated videos, where the minimum video length is 100s. To the best of our knowledge, the method to aggregate the annotations as well as the inter-rater agreement are not reported. We only employ the train set, since there are no publicly available annotations for the test partition.
\alignedsubsection{Extracted Features}
% To detect hotspots in videos, we use a ranking-based approach, where we rank 5s-long non-overlapping clips, following the annotation process of the CoView dataset. To compute the ranking, we extract the following features from the videos:
We extracted the following features from the CoView dataset.\\
% \begin{itemize}
\textbf{GoogLeNet} To have frame-wise representations of objects in images, we extract 1024-dimensional embeddings from a pre-trained GoogLeNet \cite{szegedy2015going} model.\\
\textbf{Places365} To represent scene understanding for unstructured videos, we obtain 2048-dimensional embeddings from a ResNet50 \cite{he2016deep} model pre-trained on the Places2 database \cite{zhou2017places}. We posit that videos with scenic views would be of high interest according to human judgment, which motivated including these embeddings into the model. \\
\textbf{Affect} We extract 128-dimensional embeddings for arousal and valence from the videos using the multimodal model proposed in \cite{kollias2019expression}, considering that portions of video with higher demonstrations of affect will be included in highlights.\\
\textbf{Multisensory class activation maps (CAMs)} (Owens et al.\  \cite{owens2018audio}, Zhou et al.\  \cite{zhou2016learning}) We obtain $14\times 14$ foreground CAMs per frame for each video, following the same procedure as \cite{owens2018audio} to obtain CAM visualizations but without up-sampling the actual frame size.\ We assume that interesting portions of videos contain actions where audio and video are correlated.\\
\textbf{Faces} Following Gygli et al.\ in \cite{gygli2014creating}, we use a face detection module to detect faces in videos. As a feature, we use the area of the face detected with highest confidence (and $0$ area in frames with no face detected).\ We assume that interesting portions of videos are likely to have people in them.\\
\textbf{MFCCs} We obtain 13-dimensional MFCCs from the audio sub-sampled to 16kHz, with a window length of 25ms and a window step of (0.25/29.97)s such that we obtain 4 samples per video frame. The audio features could be helpful in identifying attributes such as presence of speech, music and other acoustic sources \cite{6560388}. \\
% \end{itemize}
We converted all videos to 29.97 frames per second prior to feature extraction.\footnote{The 29.97 fps for feature extraction originates from the multisensory \cite{owens2018audio} model, which expects that sampling rate.} After extraction, all features are then upsampled to 30 fps before training the models which allows the feature frames to be aligned with the CoView labels.\ The input to the model is all features corresponding to a 5 second labeled video segment.

\begin{figure}[htp]
\centering
    \includegraphics[width=\columnwidth]{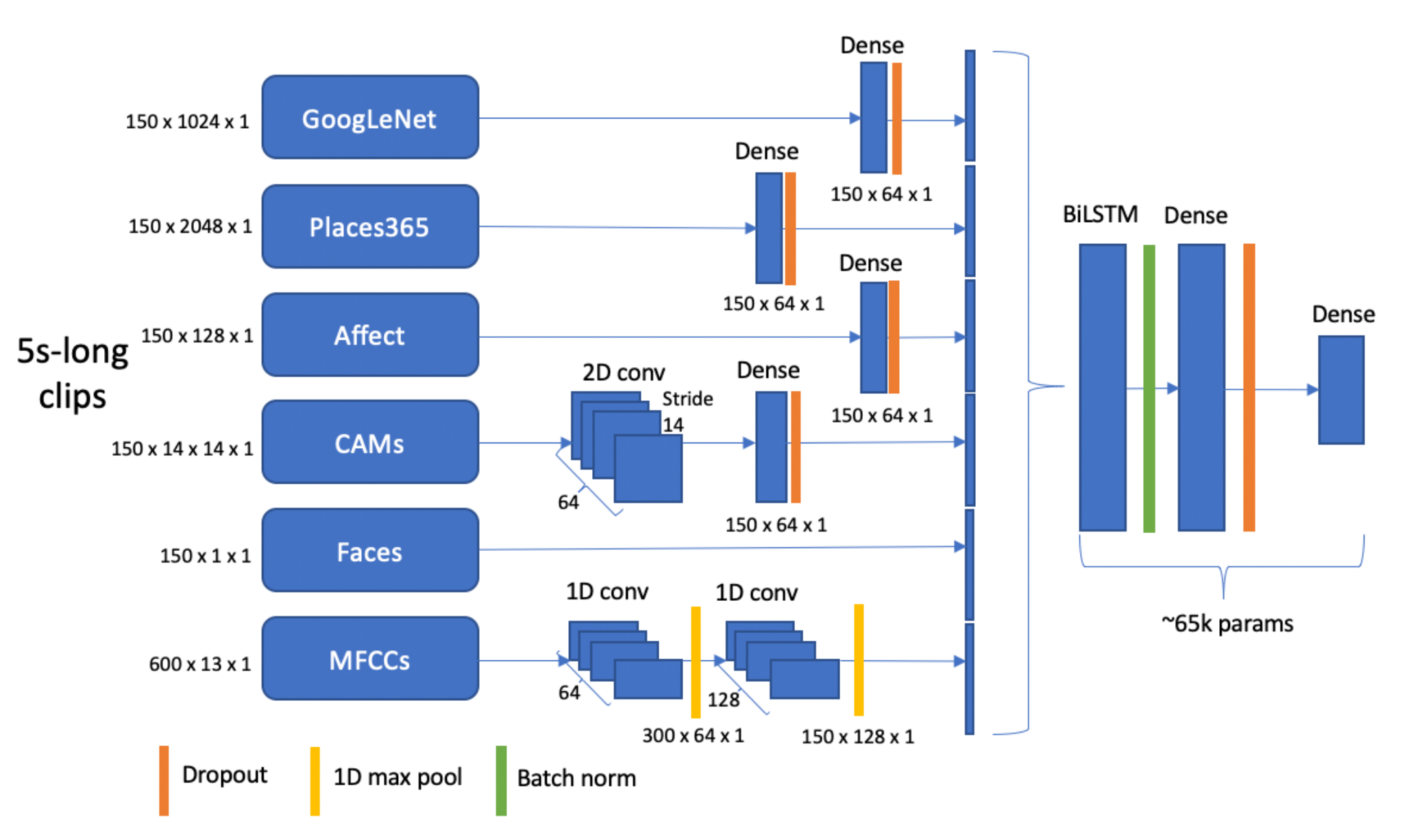}
    \vspace{-7mm}
    \caption{Supervised model architecture.}
    \label{fig:model}
    \vspace{-7mm}
\end{figure}

\alignedsubsection{Model}
We propose a modular supervised model for the prediction of summarization labels. \autoref{fig:model} shows the architecture: we employ different pre-processing schemes for different modalities.\ We then concatenate all the intermediate representations to get a single embedding vector before a bidirectional LSTM layer. We now describe each pre-processing module:\\
% \begin{itemize}
\textbf{GoogLeNet, Places365 and Affect embeddings} We use a dense layer for each embedding vector to reduce their dimension to 64.\\
\textbf{Multisensory CAMs} We employ 64 convolutional filters of size $14\times 14$ per frame, whose output is passed to a dense layer for a final dimension of 64.\\
\textbf{Faces} Representation is directly concatenated with the other embeddings.\\
\textbf{MFCCs} We employ 2 convolutional layers (in time). The first one with 64 filters and the second one with 128 filters.  Each convolutional layer is followed by a 1-dimensional, stride-2 max pool, for a final dimension of 128.\\
% \end{itemize}
We use a dropout of 0.5 after each dense layer, and both recurrent and input dropout in the bidirectional LSTM layer. The LSTM and dense layers have 20 hidden units each, and the complete model has 314k params without considering the networks from which features have been extracted.
% Ranking loss
%For experiments with ranking loss, we build a siamese architecture based on the model described above, where both sub-networks are identical and share all parameters.
\vspace{0mm}
\alignedsubsection{Loss functions}
L1 or L2 loss functions only penalize for the point-wise differences between the model output and the labels, without considering relative scores between different segments.\ Therefore, we train with two loss functions: Canonical Correlation Coefficient (CCC) loss, and ranking loss.\ Furthermore, our experiments with L2 loss also showed inferior performance when compared with CCC loss.\\
Each training example to the model represents a 5-second long clip, with the corresponding summarization score from the CoView dataset.  \\
We use the CCC loss because it penalizes both different ordinal variations in the sequences as well as differences in their means. Given inputs $x$, the labels $y$ and the output from the model $f(x)$, CCC is defined as:
\vspace{-3mm}
\begin{equation}
    \rho_c(y,f(x)) = \frac{2\sigma_y\sigma_{f(x)}}{\sigma_y^2 + \sigma_{f(x)}^2 + (\mu_y - \mu_{f(x)})^2} \in [0,1],
        \vspace{-3mm}
\end{equation}
where $\mu$ and $\sigma$ are the mean and standard deviation for each sequence.\ To convert this correlation to a loss, we use $\ell(y, f(x)) = 1 - \rho_c(y,f(x))$ as the loss.\\
% Ranking loss
Ranking loss allows the model to learn how to assign higher scores to more interesting clips from the CoView labels perspective.\ To train with margin ranking loss as defined by Yao et al. in \cite{yao7780481highlight}, we sample the set $\mathcal{P}$ of clips pairs $(h_i, l_i)$ such that the CoView label value for one clip in the pair is strictly higher than the other clip's label value. Across all pairs, the margin ranking loss is defined as:
\vspace{-1mm}
\begin{equation}
\sum\limits_{(h_i, l_i) \in \mathcal{P}} \max (0, 1 - f(h_i) + f(l_i)),
    \vspace{-3mm}
\end{equation}
where $f(\cdot)$ denotes the output score of the model.

\alignedsubsection{Training}
We train the network using 5s-long segments. We use Adam to train, with a learning rate of 0.001, $\beta_1 = 0.9$, $\beta_2 = 0.999$, and $\varepsilon=1\times10^{-7}$.\ For training with CCC loss, we use a batch size of 300, and train over 20 epochs, after which we pick the best performing model over the validation set, over all epochs.\ For training with ranking loss, we use a batch size of 64, and train over 10 epochs, and the best model is selected the same way.

\alignedsubsection{Evaluation metrics}
We report the mean of NDCG and Precision at $k$ scores across all videos in the dataset.
\vspace{-1mm}
\alignedsubsubsection{NDCG}
Since we are computing rankings, we evaluate our methods using the Normalized Discriminative Cumulative Gain (NDCG) as described by J\"arvelin et al.\ in \cite{jarvelin2002cumulated}, which yields a value related to the quality of an estimated ranking.\ If we assume that each ranked item has an estimated relevance score $\hat{r}_i$ (output of the model), then the Discriminative Cumulative Gain (DCG) over the top-$k$ items of the estimated ranking is defined as:
\vspace{-3mm}
\begin{equation}
    DCG_k = \sum_{i=1}^k \frac{\hat{r}_i}{\log_2(i+1)},\label{eq:DCG}
        \vspace{-1mm}
\end{equation}
where the $\log_2$-based denominator is a penalization term for the position of item $i$ in the ranking.\ If we also know the ground truth relevances $r_i$, we can compute the \textit{ideal} DCG (IDCG) by replacing the $\hat{r}_i$ with $r_i$ in \autoref{eq:DCG}. Then, the Normalized DCG is simply:
\vspace{-2mm}
\begin{equation}
    NDCG_k = \frac{DCG_k}{IDCG_k} \in [0,1],
    \vspace{-3mm}
\end{equation}
where a value of 1 implies perfect ranking.
\vspace{-1mm}
\alignedsubsubsection{Precision at $k$ (P@$k$)}
We use the Precision at $k$ (P@$k$) score to evaluate how many of the top-$k$ highlights	 we have been able to detect, based on the ranking method described above. The precision score is defined as ratio of true positives over all positives (TP: True positive, FP: False positive):
\vspace{-3mm}
\begin{equation}
    \text{Precision} = \frac{\text{TP}}{\text{TP} + \text{FP}},
\end{equation}
or the number of selected items that are relevant. To compute P@$k$, we sort the labels and predictions in descending order, and threshold at the top-$k$.

\vspace{-3mm}
\alignedsection{Experiments}
We run three different studies to understand the contribution of different modalities, as well as the ability to transfer the learning from unstructured videos to videos within a given topic.

\alignedsubsection{Experiments on CoView data using summarization labels}
For both experiments, we run 5-fold cross-validation for each single feature or set of features and report cross-validation performance.
\vspace{-2mm}
\alignedsubsubsection{Relevance of single features}
To understand the contribution of single features, we train the model from \autoref{fig:model} with only one feature at a time. Since the sizes of the embeddings are different for some features, we keep the number of parameters of the layers after concatenation close to 15k. For MFCCs, we reduce the number of LSTM and hidden layer units from 20 to 13, and for faces, from 20 to 4.
\vspace{-2mm}
\alignedsubsubsection{Ablation experiments}
To understand the impact of each feature in the full model, we run an ablation study, where we leave one feature out at a time from the model. Again, since the number of trainable parameters change, we adjust the number of hidden units to have approximately 65k parameters for each experiment.\\
% Ranking loss
We run ablation experiments with both CCC and ranking loss functions.\ Following Yao et al.\ \cite{yao7780481highlight}, for experiments with ranking loss we build a siamese architecture based on the model described above, where both sub-networks are identical and share all parameters. For evaluation, we run every clip through a single sub-network to produce one score per clip. Further evaluation process is identical to CCC evaluation.
% Ranking loss
\vspace{-2mm}
\alignedsubsubsection{VASNet experiment}
We compare our model with the VASNet model architecture proposed by Fajtl, Jiri et al.\ in \cite{fajtl2018summarizing}.\ Since this model was trained on frame-wise labels, and CoView only provides one label for a 5-second segment, we used the same score for all frames in the 5-second window for training.\ During inference, we computed the score of the 5-second segment as the mean of the score for each frame in the segment.\ VASNet was trained on GoogLeNet embeddings \cite{szegedy2015going}, so we find it directly comparable to our model trained on the same single feature (GoogLeNet column in \autoref{tab:single-features}).

\vspace{-1mm}
\alignedsubsectionsubsubsection{Experiments on CoView with highlight labels}{Annotations for highlight detection}
\vspace{-1mm}
For highlight annotations, we took a subset of 30 videos of the train dataset and collected annotations for the same 5s-long segments. We preferred lightly edited videos containing social interactions. Each video was annotated by three annotators.\ The annotators were instructed to watch the videos from beginning to end, and to annotate each 5s-long segment using the following instructions: ``\textit{Please rate each clip using the scale below to label the segments of the video that contain highlights, defined as an outstanding part of a video over time. Note that a video may have no highlights.}'', where the scale was: (1) \textit{Definitely not a highlight}, (2) \textit{Not a highlight}, (3) \textit{Maybe a highlight}, (4) \textit{One of the highlights}, (5) \textit{Definitely a highlight}. As preprocessing of the highlight annotations, we scale the annotated values between 1 and 5 to the interval $[0, 2]$.
\vspace{-1mm}
\alignedsubsubsection{Fine-tuning}
To fine-tune the model, we train the full model on a single fold with CoView labels.\ We then fine-tune on the 30 videos with highlight scores using 6-fold cross validation, with 25 videos for training and 5 for evaluation in each fold, with a batch size of 64 for 40 epochs, with a learning rate of \num{1e-4}. The annotations obtained from the annotators are averaged to get the labels used for training.

\vspace{-1mm}
\alignedsection{Results and Analysis}
\begin{table*}
\centering \resizebox{1.0\textwidth}{!}{\begin{minipage}{\textwidth} \centering
    \small{\caption{Single feature NDCG and precision with  CCC loss. Highest scores in \textbf{bold}, lowest scores in \textit{italics}. VASNet \cite{fajtl2018summarizing} for comparison. \label{tab:single-features} }}
    \vspace{-3mm}
    \footnotesize
    \begin{tabular}{lrcccccccc}
        \toprule
            \textbf{Metric} & $\bf{k}$ & \multicolumn{6}{c}{\textbf{Features}} & \textbf{Chance} & \textbf{VASNet} \\
        \midrule
                                            &  & \textbf{GoogLeNet} & \textbf{Places365} & \textbf{Affect} & \textbf{MFCCs} & \textbf{CAMs} & \textbf{Faces} & & \\
        \cmidrule{3-8}
        \multirow{ 2}{*}{\textbf{NDCG@$k$}} &  5 & \bf{0.763} & 0.758 & 0.743 & 0.723 & 0.726 & \textit{0.720} & 0.717 & 0.757 \\
                                          & 10 & \bf{0.802} & 0.799 & 0.783 & 0.765 & 0.769 & \textit{0.763} & 0.756 & 0.796 \\
        \midrule
        \multirow{ 2}{*}{\textbf{P@$k$}}  &  5 & \bf{0.170} & 0.163 & 0.127 & 0.135 & 0.128 & \textit{0.126} & 0.129 & 0.165 \\
                                          & 10 & \bf{0.309} & 0.306 & 0.258 & 0.262 & \textit{0.257} & 0.260 & 0.251 & 0.293 \\
        \bottomrule
    \end{tabular}
    \vspace{-3mm}  \end{minipage}}
\end{table*}
% Ranking loss
\begin{table*}
\centering  \resizebox{1.0\textwidth}{!}{\begin{minipage}{\textwidth} \centering
    \small{\caption{Leave-one-feature-out NDCG and precision using CCC loss with best models trained with Ranking Loss (RL) for comparison.\label{tab:ablation}}}
    \vspace{-3mm}
    \footnotesize
    \begin{tabular}{lrcccccccccc}
        \toprule
            \textbf{Metric} & $\bf{k}$ & \multicolumn{6}{c}{\textbf{Features CCC}} & \textbf{All CCC} & \textbf{All RL} & \textbf{RL} & \textbf{RL}\\
        \midrule
                                            &  & \textbf{No} & \textbf{No} & \textbf{No} & \textbf{No} & \textbf{No} & \textbf{No} & & &  \textbf{No} &  \textbf{No} \\
                                            &  & \textbf{GoogLeNet} & \textbf{Places365} & \textbf{Affect} & \textbf{MFCCs} & \textbf{CAMs} & \textbf{Faces} & & & \textbf{Places} & \textbf{CAMs}\\
        \cmidrule{3-8}
        \multirow{ 2}{*}{\textbf{NDCG@$k$}} &  5 & 0.764 & 0.764 & \bf{0.766} & 0.763 & \bf{0.766} & 0.765 & 0.765 & 0.729 & 0.723 & 0.715 \\
                                          & 10 & 0.804 & 0.802 & 0.804 & 0.804 & \bf{0.806} & 0.804 & 0.805 & 0.772 & 0.765 &  0.763 \\
        \midrule
        \multirow{ 2}{*}{\textbf{P@$k$}}  &  5 & 0.169 & 0.177 & \bf{0.178}  & 0.171 & 0.174 & 0.174 & 0.170 & 0.140 & 0.130 & 0.120 \\
                                          & 10 & 0.312 & 0.307 & 0.307 & 0.312 & \bf{0.316} & 0.311 & 0.312 & 0.265 & 0.256 & 0.263 \\
        \bottomrule
    \end{tabular}
    %}
    \vspace{-4mm}  \end{minipage}}
\end{table*}
\vspace{3mm}
\begin{figure}[t]
    \vspace{-3mm}
    \centering
   \resizebox{\linewidth}{!}
    {\input{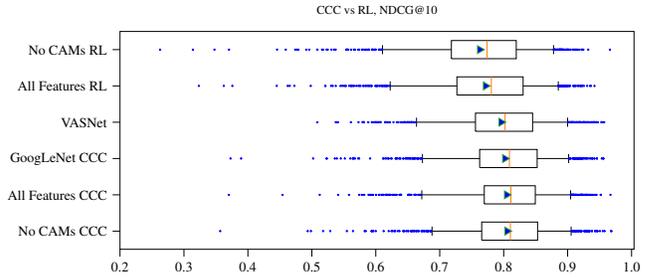}}
    \vspace{-6mm}
    \caption{Box plots for NDCG@10 for best performing models. Outer whiskers set to 5th and 95th percentiles. Triangles are at mean values, vertical lines are at medians and stars are outliers.}
    \vspace{-8mm}
    \label{fig:boxplots}
\end{figure}

\alignedsubsectionsubsubsection{Experiments on CoView data and labels}{Relevance of single features}
\vspace{-1mm}
We present the results using single features in \autoref{tab:single-features}, where the ``Chance'' column shows the chance values for all metrics. The results shown are the performance over 5-fold cross validation. For this task, we observe the best performance from the GoogLeNet embeddings (trained on ImageNet) across all metrics, followed by the Places365 embeddings (trained on Places2), while the least informative features are the proportion of the faces in the frames. For P@5, faces, affect, and multisensory CAMs are below chance.\\

\begin{figure}
    \centering
    % This file was created by tikzplotlib v0.9.3.
\begin{tikzpicture}

\definecolor{color0}{rgb}{0.12156862745098,0.466666666666667,0.705882352941177}

\begin{axis}[
align=center,
width=0.33\columnwidth,
height=0.14\columnwidth,
scale only axis,
tick align=outside,
tick pos=left,
x grid style={white!69.0196078431373!black},
xlabel={\scriptsize{Cronbach's $\alpha$}},
xmajorgrids,
xmin=0, xmax=1,
xtick style={color=black},
xticklabels={{\scriptsize 0.0}, {\scriptsize 0.0}, {\scriptsize 0.2}, {\scriptsize 0.4},{\scriptsize 0.6},{\scriptsize 0.8},{\scriptsize 1.0}},
y grid style={white!69.0196078431373!black},
y label style={font=\scriptsize},
ylabel={Number of\\ videos},
yticklabels={{\scriptsize 0}, {\scriptsize 0}, {\scriptsize 2}, {\scriptsize 4},{\scriptsize 6}},
ymajorgrids,
ymin=0, ymax=7.35,
ytick style={color=black}
]
\draw[draw=none,fill=color0] (axis cs:0.01,0) rectangle (axis cs:0.099,2);
\draw[draw=none,fill=color0] (axis cs:0.099,0) rectangle (axis cs:0.188,0);
\draw[draw=none,fill=color0] (axis cs:0.188,0) rectangle (axis cs:0.277,1);
\draw[draw=none,fill=color0] (axis cs:0.277,0) rectangle (axis cs:0.366,3);
\draw[draw=none,fill=color0] (axis cs:0.366,0) rectangle (axis cs:0.455,3);
\draw[draw=none,fill=color0] (axis cs:0.455,0) rectangle (axis cs:0.544,7);
\draw[draw=none,fill=color0] (axis cs:0.544,0) rectangle (axis cs:0.633,3);
\draw[draw=none,fill=color0] (axis cs:0.633,0) rectangle (axis cs:0.722,6);
\draw[draw=none,fill=color0] (axis cs:0.722,0) rectangle (axis cs:0.811,4);
\draw[draw=none,fill=color0] (axis cs:0.811,0) rectangle (axis cs:0.9,1);
\end{axis}

\end{tikzpicture}
    % This file was created by tikzplotlib v0.9.3.
\begin{tikzpicture}

\definecolor{color0}{rgb}{0.12156862745098,0.466666666666667,0.705882352941177}

\begin{axis}[
align=center,
width=0.33\columnwidth,
height=0.14\columnwidth,
scale only axis,
tick align=outside,
tick pos=left,
x grid style={white!69.0196078431373!black},
xlabel={\scriptsize{Kendall's rank correlation}},
xmajorgrids,
xmin=-0.3, xmax=0.7,
xtick style={color=black},
xticklabels={{\scriptsize -0.3}, {\scriptsize -0.2}, {\scriptsize 0}, {\scriptsize 0.2},{\scriptsize 0.4}, {\scriptsize 0.6}},
y grid style={white!69.0196078431373!black},
y label style={font=\scriptsize},
ylabel={Number of\\videos},
yticklabels={{\scriptsize 0}, {\scriptsize 0}, {\scriptsize 2}, {\scriptsize 4},{\scriptsize 6}},
ymajorgrids,
ymin=0, ymax=7.35,
ytick style={color=black}
]
\draw[draw=none,fill=color0] (axis cs:-0.229253256829021,0) rectangle (axis cs:-0.142287237086068,2);
\draw[draw=none,fill=color0] (axis cs:-0.142287237086068,0) rectangle (axis cs:-0.0553212173431152,0);
\draw[draw=none,fill=color0] (axis cs:-0.0553212173431152,0) rectangle (axis cs:0.0316448023998377,1);
\draw[draw=none,fill=color0] (axis cs:0.0316448023998377,0) rectangle (axis cs:0.118610822142791,4);
\draw[draw=none,fill=color0] (axis cs:0.118610822142791,0) rectangle (axis cs:0.205576841885743,7);
\draw[draw=none,fill=color0] (axis cs:0.205576841885743,0) rectangle (axis cs:0.292542861628696,2);
\draw[draw=none,fill=color0] (axis cs:0.292542861628696,0) rectangle (axis cs:0.379508881371649,6);
\draw[draw=none,fill=color0] (axis cs:0.379508881371649,0) rectangle (axis cs:0.466474901114602,5);
\draw[draw=none,fill=color0] (axis cs:0.466474901114602,0) rectangle (axis cs:0.553440920857555,1);
\draw[draw=none,fill=color0] (axis cs:0.553440920857555,0) rectangle (axis cs:0.640406940600507,2);
\end{axis}

\end{tikzpicture}
    \vspace{-4mm}
    \caption{Annotator agreements and correlations with CoView labels. (Left) Histogram with Cronbach's $\alpha$ for the 30 annotated videos. (Right) Histogram with Kendall's rank correlation between aggregated annotations and  CoView labels.}
    \vspace{-3mm}
    \label{fig:highlights}
\end{figure}
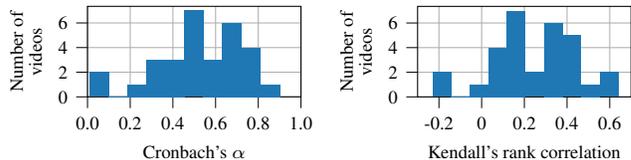

\vspace{-5mm}
\alignedsubsubsection{Ablation experiments}
For training with CCC loss, we present the results using different subsets of features in \autoref{tab:ablation}, where the ``All CCC'' column shows the results when training with all features.\ For this task, we observe that including most features in the model beats chance, and removing multisensory CAMs produces the best cross validation results when we train with CCC loss. We also observe that except for the model without multisensory CAMs, most models have a performance similar to using GoogLeNet embeddings alone. We show the NDCG@10 box-plots for the best performing models in \autoref{fig:boxplots}. The figure shows that although the mean and median NDCG values of the model trained with GoogleNet alone are similar to the proposed model, the first quartile is higher for our proposed model.\\
We also observe that results from training with CCC loss are always superior to results from training with ranking loss for all ablation experiments performed. \autoref{fig:boxplots} reflects this superiority for best performing models within each loss as well. Best model trained with ranking loss is the one trained on all features.\
 \vspace{-1mm}
\alignedsubsubsection{VASNet experiment}
First and third quartile, mean and median values for VASNet \cite{fajtl2018summarizing} model are shifted to the left in \autoref{fig:boxplots} in comparison with models trained with CCC loss, which indicates that all three CCC models improve over VASNet \cite{fajtl2018summarizing} results by a noticable margin.\\
\vspace{-5mm}
\alignedsubsectionsubsubsection{Experiments on CoView with highlight labels}{Highlight Annotations}
\vspace{-1mm}
\autoref{fig:highlights} shows the inter-rater agreement (Cronbach's $\alpha$) among annotators for all videos, with a mean of 0.524, max of 0.9, and min of 0.01.\ This figure shows the subjectivity of the task:\ even though there are videos for which the annotators agreed on the highlights, there are a large number of videos for which the agreement is low. We also show Kendall's rank correlation with CoView labels, with a mean of 0.244, max value of 0.640, and min value of -0.230. This shows the difference in the tasks, and invites the question of whether we can transfer the knowledge from one set of labels to the other.
\vspace{-1mm}
\alignedsubsubsection{Fine-tuning results}
We present the fine-tuning results in \autoref{tab:fine-tuning}. The baseline model is trained on 30 videos. For fine-tuning, we initialize the model weights learned by training the model on summarization labels with all the features (All CCC in \autoref{tab:ablation}).\ The results show that fine-tuning can effectively transfer knowledge from the summarization task to the highlight task, with the fine-tuned model demonstrating an improvement over all the metrics. Both models, however, have a lower than chance NDCG score. NDCG depends not just on the relative ranking between segments, but also on the actual scores assigned to each segment. The precision values show that the transfer learned model ranks the segment better, but the NDCG metric show that the scores are not calibrated well. We will focus on this in our future work.

\vspace{-1mm}
\alignedsection{Discussion}
\vspace{-0mm}
Our results suggest that most of the information is obtained from the GoogLeNet embeddings.\ In CoView, there are many edited videos with background music,\ which is a potential reason why multisensory CAMs as well as affect, which were trained on datasets with no background music, affect model performance. \\
When we select videos with social settings as a target set for the model fine-tuning, we see increases in P@5 and P@10, which are reflected in the selection of high-quality segments. During qualitative analysis, we observed that visually appealing clips with objects and/or people and positive emotional moments such as cheering or positive responses are preferred by the model.

\vspace{-1mm}
\begin{table}[t]
\centering
    \caption{Fine-tuning results on highlight labels.} \centering
    \vspace{-3mm}
    \footnotesize
    \begin{tabular}{lrccc}
        \toprule
            \textbf{Metric} & $\bf{k}$ & \textbf{Chance} & \textbf{Baseline} & \makecell{\textbf{Fine tuning}}\\
        \midrule
            \multirow{ 2}{*}{\textbf{NDCG@$k$}} &  5 & 0.677 & 0.581 & 0.617 \\
                                                & 10 & 0.721 & 0.612 & 0.665 \\
            \multirow{ 2}{*}{\textbf{P@$k$}}    &  5 & 0.104 & 0.133 & 0.160 \\
                                                & 10 & 0.211 & 0.200 & 0.263 \\
        \bottomrule
    \end{tabular}
    \vspace{-6mm}
    \label{tab:fine-tuning}
\end{table}

% \vspace{-2mm}
\alignedsection{Conclusions}
\vspace{-0mm}
In this work, we present an audiovisual model to detect highlights in videos. We use video summarization labels, and we find that visual features carry most of the information for highlight detection in CoView dataset. Using a subset of the proposed features, a combination of both audio and visual features, allows to obtain better performance in the selected metrics. Similar to other works that combine lexical information with audio and video  \cite{makhervaks2020combining}, as future work we propose to explore the video transcription text as additional feature to improve performance on the task. We also propose to explore end-to-end model training on a larger dataset to adapt embeddings specifically for the highlight detection task.

\alignedsection{Acknowledgements}
We thank Arpita Shah and Srinivas Parthasarathy for their help and support.

\vfill\pagebreak

% \bibliographystyle{IEEEbib}
% \bibliography{references}

\end{document}